\documentclass[letterpaper, 10 pt, conference]{ieeeconf}  

\IEEEoverridecommandlockouts                              

\usepackage{caption}
\usepackage[skins]{tcolorbox}  
\usepackage{subcaption}
\usepackage{multicol}
\usepackage{mathtools}
\usepackage{graphics} 
\usepackage{epsfig} 
\usepackage{mathptmx} 
\usepackage{times} 
\usepackage{amsmath} 
\usepackage{amssymb}  
\usepackage{graphicx}
\usepackage{epstopdf}
\usepackage{pdfpages}
\usepackage{cite}
\usepackage{amsbsy}
\usepackage{bm}
\usepackage{url}
\usepackage{accents}

\pdfoutput=1
\usepackage{hyperref}
\hypersetup{
    colorlinks=true,
    linkcolor=black,
    citecolor=black,
    filecolor=black,
    urlcolor=black,
}

\def\horizontaldistance{\kern2pt}

\title{\LARGE \bf
Surena-V: A Humanoid Robot for Human-Robot Collaboration with Optimization-based Control Architecture
}

\author{Mohammad Ali Bazrafshani$^{1}$$^\dagger$, Aghil Yousefi-Koma$^{1}$$^\dagger$, Amin Amani$^{1}$$^\dagger$, Behnam Maleki$^{1}$$^\dagger$, Shahab Batmani$^{1}$$^\dagger$\\
Arezoo Dehestani Ardakani$^{1}$$^\dagger$, Sajedeh Taheri$^{1}$$^\dagger$, Parsa Yazdankhah$^{1}$$^\dagger$, Mahdi Nozari$^{1}$$^\dagger$, Amin Mozayyan$^{1}$$^\dagger$\\
Alireza Naeini$^{1}$$^\dagger$,  Milad Shafiee$^{2}$$^\dagger$ and Amirhosein Vedadi$^{1}$$^\dagger$
\thanks{$\dagger$ All authors equally contributed to the paper.}
\thanks{$^{1}$Center of Advanced Systems and Technologies (CAST) School of
Mechanical Engineering, College of Engineering, University of Tehran,
Tehran, Iran.
        {\tt\small aykoma@ut.ac.ir}}%
\thanks{$^{2}$École polytechnique fédérale de Lausanne (EPFL)}
       }

\makeatletter
\newcommand*{\rom}[1]{\expandafter\@slowromancap\romannumeral #1@}
\makeatother
\begin{document}

\maketitle
\thispagestyle{empty}
\pagestyle{empty}

\begin{abstract}
This paper presents Surena-V, a humanoid robot designed to enhance human-robot collaboration capabilities. The robot features a range of sensors, including barometric tactile sensors in its hands, to facilitate precise environmental interaction. This is demonstrated through an experiment showcasing the robot's ability to control a medical needle's movement through soft material. Surena-V's operational framework emphasizes stability and collaboration, employing various optimization-based control strategies such as Zero Moment Point (ZMP) modification through upper body movement and stepping. Notably, the robot's interaction with the environment is improved by detecting and interpreting external forces at their point of effect, allowing for more agile responses compared to methods that control overall balance based on external forces. The efficacy of this architecture is substantiated through an experiment illustrating the robot's collaboration with a human in moving a bar. This work contributes to the field of humanoid robotics by presenting a comprehensive system design and control architecture focused on human-robot collaboration and environmental adaptability.

\end{abstract}

\section{INTRODUCTION}
Humanoid robots have emerged as versatile and capable machines designed to mimic human-like characteristics and movements. In recent years, the landscape of humanoid robotics has witnessed significant strides, propelled by advancements in technology. This progress has led humanoid robots to the forefront of real-life applications, where they are increasingly deployed to collaborate and interact more effectively with humans and their environments. This growing demand for versatile robotic solutions has spurred the development of commercial general-purpose humanoid robots that exhibit cooperative capabilities. Companies such as Tesla\cite{Tesla}, Figure \cite{Figure}, Unitree \cite{Unitree_H1}, and numerous others have unveiled their humanoid platforms.

\begin{figure}[ht]
\centering
\includegraphics[width=0.38\textwidth, trim={3900 400 3000 400}]{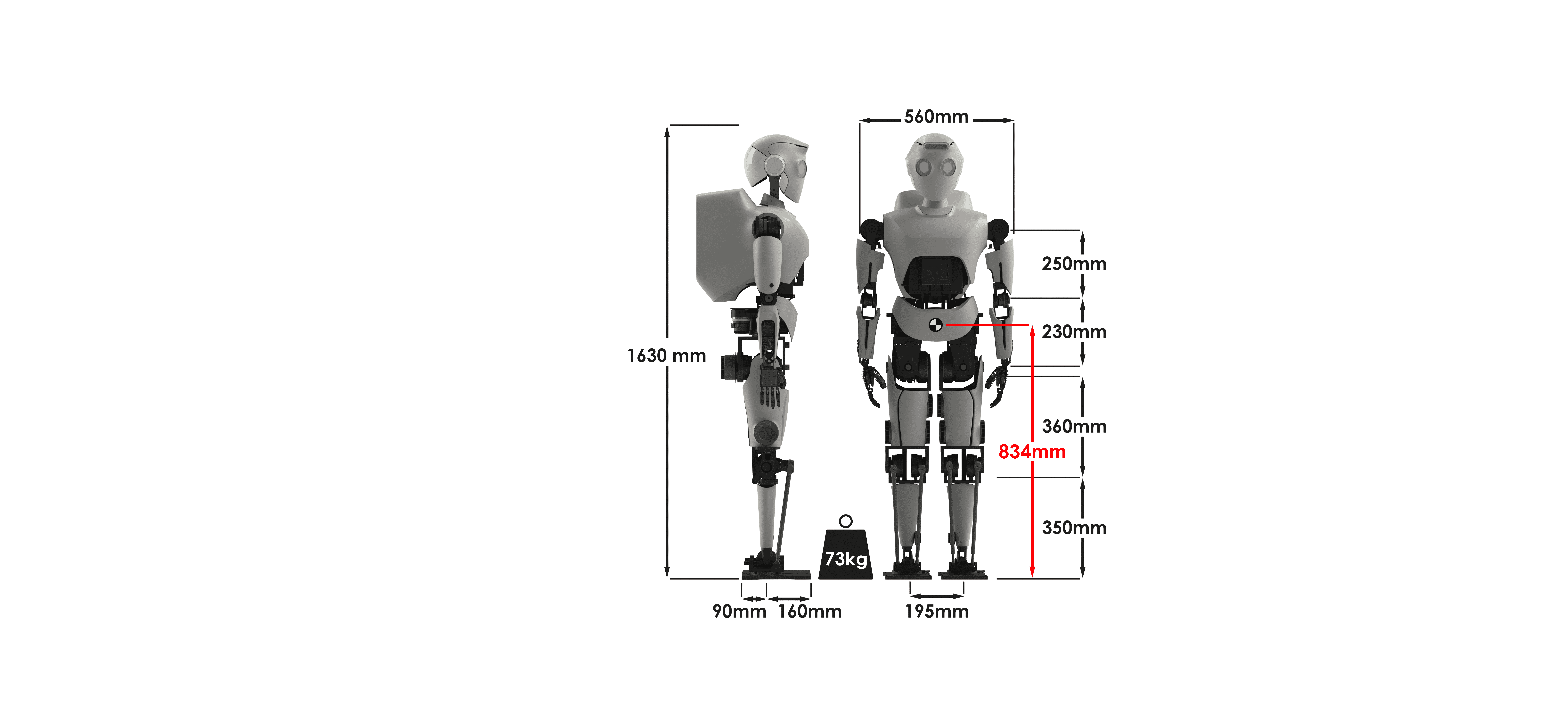}
\caption{Surena-V Humanoid Robot Specification}
      \label{fig:dimensions}
\end{figure}

In contemporary humanoid robot design, there is a notable emphasis on utilizing high-torque and compact actuators, along with reducing the inertia of lower limbs. This approach aims to enhance the agility and performance of humanoid robots. An exemplary platform showcasing these design principles is the MIT humanoid robot, which demonstrates the capability to achieve a 30 cm jump utilizing its powerful motors \cite{mitHumanoid}. Additionally, Gazelle, another innovative humanoid robot, has implemented parallel mechanisms in its ankle joint to reduce limb inertia, thereby improving its efficiency in locomotion tasks \cite{gazelle}. By integrating mechanisms such as parallel structures or compliant elements, humanoid robots can achieve enhanced rigidity and balance in their movements.

In the development of a humanoid robot, it is essential not only to focus on optimizing its performance but also to consider its capacity for effective collaboration and interaction with humans. This necessitates the incorporation of sensors and controllers specifically designed to facilitate seamless cooperation between the robot and its human counterparts.

When robot interaction is discussed, some information about the interaction properties can be approximated using the robot’s intrinsic sensors such as joint position and torque sensors. However, extrinsic (tactile) sensors give much more precise information about the interaction. These sensors which are usually used in arrays, are required for the robot to explore, manipulate, and respond to its environment. Different types of tactile sensations, such as bending, stretching, pressure, and temperature, help humans gain a sense of an object they interact with. Robots achieve this perception through different sensors embedded inside them. Different sensing mechanisms are used to create flexible tactile sensors suitable for robotic applications such as robotic arms, prosthetic hands, and humanoids. 

Van den Heever et al. \cite{tactile1} introduced a tactile sensor array utilizing force sensing resistors (FSRs) with a super-resolution algorithm to enhance spatial resolution, similar to image processing techniques. Similarly, Choi et al. \cite{tactile2} presented a flexible fingertip tactile sensor using PVDF and pressure variable resistor ink for detecting normal force and incipient slip, along with an experimentally validated tactile sensing system. Additionally, Epstein et al. \cite{tactile3} showcased new fingertip and footpad sensors employing artificial neural networks for 3-axis force and location detection, improving robot manipulation and locomotion capabilities. Tenzer et al. \cite{tactile4} proposed a cost-effective method utilizing barometric pressure sensor chips and standard PCBs to fabricate tactile array sensors with high signal quality, housed in soft polymers. Furthermore, Bhirangi et al. \cite{tactile5} introduced ReSkin, a tactile soft sensor that leverages machine learning and magnetic sensing technologies.

In the realm of robot control, various strategies have been developed to ensure the balance of position-controlled humanoid robots. Caron et al. \cite{caron} proposed a control architecture that integrates quadratic programming-based wrench distribution with whole-body admittance control. Shaffiee et al. \cite{shafiee} implemented a unified Model Predictive Control (MPC) approach incorporating Center of Pressure (CoP) manipulation, step adjustment, and Center of Mass Projection (CMP) modulation. Jeong et al. \cite{jeong} introduced a biped walking controller that optimizes ankle, hip, and stepping strategies to enhance push recovery through an integrated optimization framework.

In this paper, we present the design and development of Surena-V, building upon our previous work with Surena-IV \cite{yousefikoma2021surena}. We introduce a hand design equipped with integrated barometric tactile sensors at the fingertips, offering low latency and good accuracy. We validate this design through an experiment where the robot manipulates a medical needle through soft material, halting its movement upon detecting material stress. Furthermore, we explore the robot's operational architecture and present a whole-body control framework tailored for collaborative tasks with humans. To improve the interaction of the robot with the environment, the external force is detected and understood at the point of effect on the robot. Unlike methods that control the overall balance of the robot under the influence of external forces, this approach provides better agility. This architecture prioritizes maintaining balance by combining diverse strategies through a quadratic programming optimization approach.

In the following, section II provides an overview of the mechanical and electrical design of the Surena-V humanoid robot. Transitioning to section III, we delve into the operational architecture and control loop of Surena-V, discussing its decision-making processes, and control strategies. Section IV presents the experiment results with analysis, and finally, section V serves as the conclusion of the paper.

\section{System Overview}

The Surena-V humanoid robot boasts 41 degrees of freedom (DoF), offering its limbs and body diverse movement capabilities. Standing 1.63 meters tall and weighing 73 kg, the robot's dimensions are further detailed in Figure 1. Also for a detailed breakdown of the actuators powering each DoF, please refer to Table I. Its core structure leverages a combination of 7075 aluminum alloy for strength and carbon fiber for weight reduction. Additionally, its outer cover utilizes 3D-printed polylactic acid (PLA) technology, which offers a balance of affordability, durability, and lightweight properties. This section delves into the details of the robot's design, encompassing the upper body, lower body, and the electronics and communication systems that coordinate its movements and operations.

\begin{table*}[t]
    \begin{center}
        \caption{Specification of Joints}
        \label{tab2}
        \begin{tabular}{cccccc}
			\textbf{Joint}	& \textbf{Motor Name}	& \textbf{Mechanism}	&\textbf{$\dot{q}_{max}$,$[Rad/Sec]$}     & \textbf{$\tau_{max}$,$[N.m]$}     & \textbf{Range [min, max]$[Degrees]$}\\
            \hline 
            \noalign{\vskip 1.5mm}
			Hip Pitch	& Maxon EC90 Flat	& HarmonicDrive	& 3.5			& 160			& [-50, +60]\\
			Hip Roll	& Maxon EC90 Flat	& HarmonicDrive	& 3.5			& 160			& [-30, +15]\\
			Hip Yaw	& Maxon EC60 Flat		& HarmonicDrive+Linkage	& 5			& 30			& [-20, +10]\\
   			Knee (Pitch)	& Maxon EC90 Flat		& HarmonicDrive+Linkage	& 3.5			& 160			& [-5, +90]\\
			Ankle Pitch	& Maxon EC90 Flat		& HarmonicDrive+Parallel Mechanism	& 4.5			& 120			& [-50, +40]\\
			Ankle Roll	& Maxon EC90 Flat		& HarmonicDrive+Parallel Mechanism& 4.5			& 90			& [-20, +20]\\
			Shoulder Pitch	& Maxon EC60 Flat		& HarmonicDrive	& 10			& 30			& [-110, +80]\\
			Shoulder Roll	& Maxon EC60 Flat		& HarmonicDrive	& 10			& 30			& [-90, -5]\\
			Shoulder Yaw	& Maxon EC45 Flat		& HarmonicDrive	& 8			& 15			& [-60, +60]\\
			Elbow (Pitch)	& Maxon EC45 Flat		& Belt+HarmonicDrive	& 6			& 20			& [-90, 0]\\
			Wrist Pitch	& 6221MG Servo		& Parallel Mechanism	& 2.5			& 2.2			& [-20, +20]\\
			Wrist Roll	& 6221MG Servo		& Parallel Mechanism	& 2.5			& 2.2			& [-20, +20]\\
			Wrist Yaw	& 6221MG Servo		& Direct Drive	& 3.5			& 2			& [-90, +90]\\
			Neck Pitch	& 6221MG Servo		& Direct Drive	& 3.5			& 2			& [-30, +20]\\
			Neck Roll	& 6221MG Servo		& Direct Drive	& 3.5			& 2			& [-30, +30]\\
			Neck Yaw	& 6221MG Servo		& Direct Drive	& 3.5			& 2			& [-90, +90]\\
			Fingers	& ZGA16		& Lead Screw+Linkage	& 0.5			& 0.5		& [0, +90]\\
        \end{tabular}
    \end{center}
\end{table*}

\subsection{Upper Body Design}

Each arm boasts seven DoF, allowing for versatile manipulation tasks. Additionally, the hands, equipped with six DoF each, further enhance the robot's grasping capabilities. Moreover, the head, possessing three DoF, and outfitted with interactive LED eyes, fosters improved interaction with humans by conveying visual expressions \cite{ROJASQUINTERO2021103834}.

To enhance the robot's resemblance to an inverted pendulum model, a commonly used control approach for humanoid robots, approximately 60 percent of its total mass is concentrated within the upper body. This mass distribution plays a crucial role in achieving balance and stability during locomotion. Figure 2 provides a detailed visualization of the various components constituting the upper body.
\subsubsection{Wrist Parallel Mechanism}
Designing robotic wrists has always been challenging due to the necessity for compact and robust designs to facilitate effective manipulation tasks.
A $2\underline{R}SS - 1U$ wrist mechanism design was proposed for our humanoid robot SURENA-V, which is lightweight, compact, and has high load capacity \cite{taheri2023design}.
As shown in Figure 2, the revolute joints are driven by two servo motors. A universal joint connects the end-effector to the motor case. Both the joint connecting the limbs to the moving plate and the joint connecting the horns to the limbs are spherical joints.
The wrist provides two DoF for  flexion/extension (F/E) and radial/ulnar deviation (R/U) movements.

The length of $horn_{i}$ and $limb_{i}$, $i = 1, 2$ are the design variables of the mechanism.
These variables are defined by minimising the general condition number (GCN) over n discrete points in the whole workspace \cite{taheri2023design}. Condition number is a performance criterion used to characterise the dexterity of the manipulator.

The optimisation problem is expressed as follows:
\begin{equation}
\begin{split}
    minimize     \hspace{15pt}   &GCN(J)=\frac{\sum_{i=1}^{n}(||J||.||J^{-1}||)}{n} \label{eq:eq9}\\
    subjected\hspace{5pt} to  \hspace{15pt} &0<horn_{i}<0.15 \hspace{30pt} i=1,2\\
    &0.4<limb_{i}<0.8 \hspace{30pt} i=1,2
\end{split}
\end{equation}
Where J is the Jacobian matrix of the wrist. The result of solving the optimisation problem is shown in Table II.

\begin{table}[]
\centering
\caption{The result of solving the wrist's optimisation problem.}
\begin{tabular}{cc}
\hline
$horn_1$  & 15  mm   \\ \hline
$horn_2$  & 15  mm   \\ \hline
$limb_1$  & 61.8 mm \\ \hline
$limb_2$  & 78.6 mm \\ \hline
\end{tabular}
\label{table:constraint}
\end{table}

\subsubsection{Hand Design}
The Surena-V's hand design prioritizes collaborative interaction with humans. To achieve this, we employed RTV-2 silicone rubber for the hand's skin. This material offers several advantages:

\begin{itemize}
  \item Compliance: It allows the hand to conform to various object surfaces, enhancing grasping reliability.
  \item Friction: The inherent friction of the material improves grip strength and object manipulation.
  \item Tactile comfort: The silicone provides a more human-like feel when touched, promoting user acceptance and reducing potential anxieties during interaction.
\end{itemize}

Each hand boasts six degrees of freedom (DoF), with individual DoF allocated to each finger except the thumb. The thumb possesses an additional DoF, granting it the critical ability to oppose (move in front of) the other fingers, enabling a wider range of grasping configurations.

High-torque motor-gearboxes drive each DoF. These motors are coupled with lead-screw mechanisms to convert rotary motion into linear displacement. The lead screw, in turn, actuates a linkage system responsible for finger articulation. A detailed illustration of this mechanism can be found in Figure 2.

To imbue the robot with a sense of touch, we implemented tactile sensors based on fluid pressure sensing. Each fingertip houses a sealed container encompassing a BMP280 pressure sensor. This design allows the robot to gauge the applied grasping force, enabling it to adapt its grip strength dynamically and fostering safe and nuanced interaction with humans during collaborative tasks.

\begin{figure*}[]
\centering
\includegraphics[height=0.6\textheight, trim={0 0 0 0}]{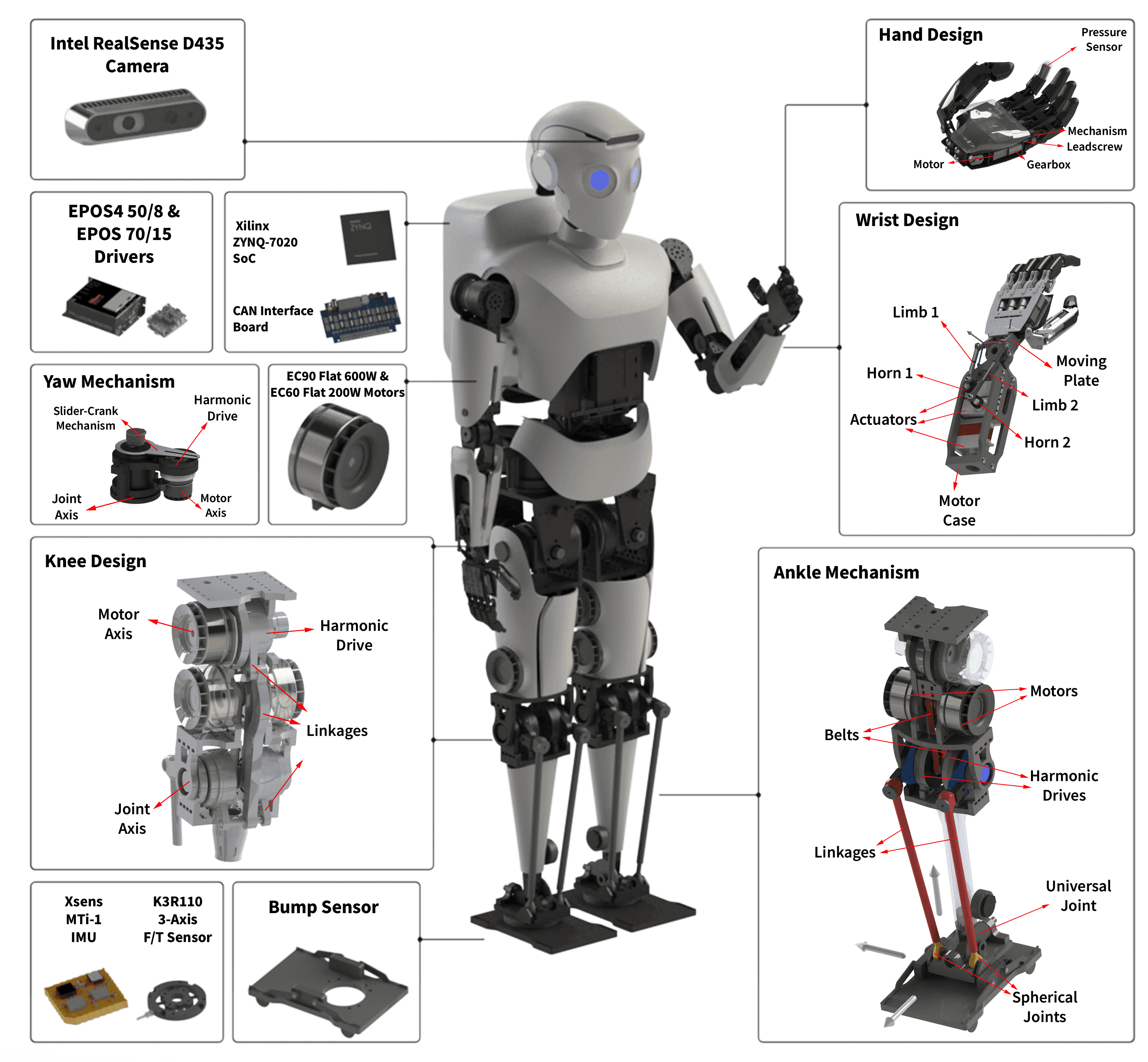}
\label{fig:design}
\caption{Comprehensive View of the Robot with Integrated Parts and Mechanisms}
\end{figure*}

\subsection{Lower Body Design}
The design of the robot's lower body focuses on achieving a balance between mobility and control for effective locomotion. All motors are strategically positioned within the robot's thighs. The motors transmit motion to the joints through linkages and mechanisms, minimizing the moving mass of the legs and enhancing maneuverability.

Each foot sole is equipped with a 3-axis FT sensor and bump sensor. FT sensor measures the forces and torques acting on the foot, providing crucial information for maintaining balance and stability during various maneuvers. Integrated bump sensors detect vertical distances to the ground within a 2 cm range.

\subsubsection{Hip Yaw Joint Mechanism}
The hip yaw joint of the robot is operated by a slider-crank mechanism, allowing for the motor to be positioned next to the joint axis, thereby enhancing the compactness of the robot's design. This mechanism is illustrated in Figure 2.
\subsubsection{Knee Joint Mechanism}
A four-bar parallel mechanism facilitates knee actuation. This design allows the knee motor to be positioned in the upper body while connecting to the joint through the linkage. Notably, the joint and motor angles are directly proportional due to the mechanism's kinematic properties.
\subsubsection{Ankle Joint Mechanism}
A parallel mechanism also drives the ankle joint. This configuration offers several advantages:
\begin{itemize}
      \item Increased Ankle Stiffness: The mechanism enhances the overall stiffness of the ankle, contributing to better support for maintaining posture.
      \item Reduced Leg Inertia: Similar to the hip-mounted actuation, the placement of the ankle motor within the leg structure further reduces leg inertia and minimizes the D'Alembert effect during leg swing, a phenomenon that can hinder dynamic movement.
\end{itemize}

The dimensions of the mechanism are carefully optimized to ensure sufficient torque generation throughout the ankle's workspace. Figure 2 illustrates this mechanism.

\begin{figure*}[]
\centering
\includegraphics[width=0.8\textwidth, trim={0 0 0 0}]{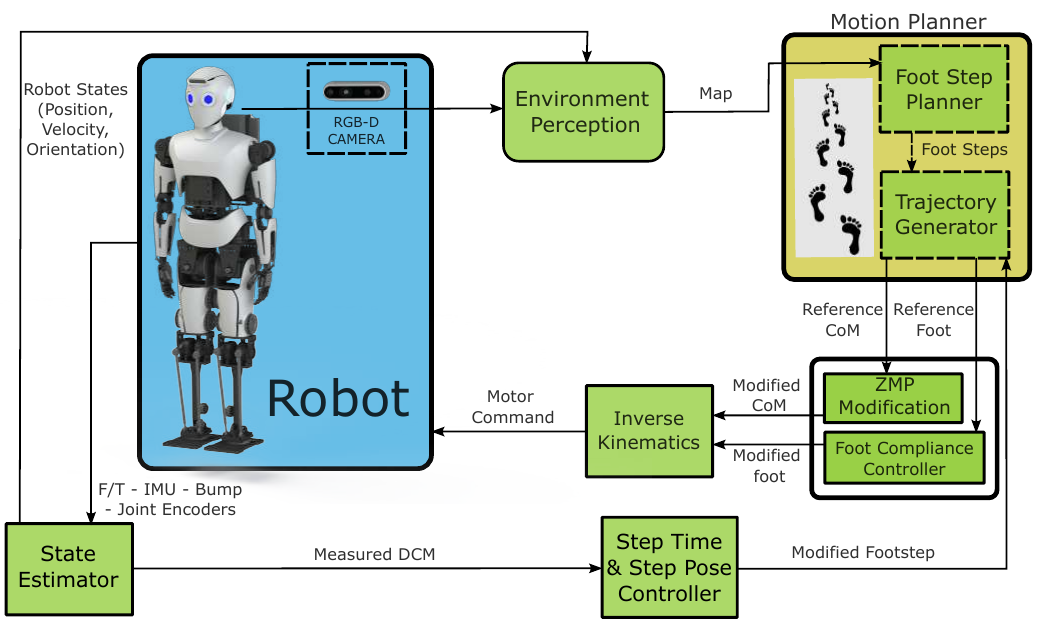}
\label{fig:architecture}
\caption{Operational Architecture of the Robot}
\end{figure*}

\subsection{Electronics and Communications}
The Surena-V's central system comprises several critical components that enable its real-time movement and intelligent behavior. An 11th Gen Intel® Core™ i7-11370H CPU serves as the robot's brain, handling high-performance tasks such as real-time movement generation and control. Also, An NVIDIA GeForce RTX 3060 GPU with 8GB VRAM empowers the robot's visual processing capabilities and supports various AI tasks. For seamless communication within the computer between various computational nodes, we employed the Robot Operating System (ROS).

To facilitate seamless communication between the diverse sensors and actuators, the robot leverages a Xilinx ZYNQ-7020 system-on-chip (SoC). This versatile board integrates dual ARM Cortex-A9 Processors and a Field-Programmable Gate Array (FPGA) which provides custom hardware acceleration for critical real-time tasks. Operating at a frequency of 200 Hz, the ZYNQ-7020 acts as a central communication hub:
\begin{itemize}
  \item Sensor Data Acquisition: It gathers data from various sensors, including force-torque sensors, IMU, and bump sensors.
  \item Data Transmission: The acquired data is transmitted to the main processor through a Gigabyte UDP protocol.
  \item Command Reception: It receives new control commands from the main processor.
  \item Motor Driver Communication: The ZYNQ-7020 communicates with individual motor drivers using a CAN bus network. This network is further segmented to optimize communication efficiency:
      \begin{itemize}
      \item Lower Body Motors: Each motor driver has a dedicated CAN bus connection.
      \item Upper Body Arms: All seven DoF motors share a single CAN bus due to their moderate data traffic.
      \item Hands: Each hand utilizes a dedicated CAN bus due to the high volume of data associated with finger position and pressure readings.
      \end{itemize}
\end{itemize}

\section{Robot Operational Architecture}

Achieving robust autonomy in complex environments for a generalized humanoid robot requires the seamless integration of various interconnected functionalities. These functionalities can be broadly categorized within the robot's operational architecture, as depicted in Figure 3.

Environment perception leverages simultaneous localization and mapping (SLAM) algorithms \cite{slam} to build and maintain a map of its surroundings, while concurrently employing object detection and tracking algorithms to identify and localize objects within the environment. Based on this information, the footstep planner generates collision-free and dynamically feasible footholds for the robot. Subsequently, these planned footholds are used to generate trajectories for the robot's foot and kinematic base movement \cite{trajectory}. The control system then executes these trajectories, employing real-time feedback mechanisms to compensate for disturbances and ensure smooth and stable motion. To achieve optimal performance, the robot continuously estimates its internal states (e.g., base position and velocity) \cite{estimation} and feeds them back into the perception and control modules, closing the loop and enabling continuous adaptation and refinement of its behavior.
\subsection{Control Strategies}
To address the inherent uncertainties associated with modeling and manufacturing imperfections, achieving robust and precise control of humanoid robots necessitates the implementation of control strategies that leverage sensor-based state estimation. In this section, we will describe these control strategies, explaining how they maintain robot balance and ensure accurate trajectory tracking.
\subsubsection{Foot Compliance Controller}
Maintaining balance and stability during legged locomotion on uneven terrain necessitates real-time adaptation of the foot's interaction with the ground. To achieve this, our humanoid robot employs a foot compliance control strategy based on previously published research \cite{compliance}. This strategy addresses two key challenges: leg length adjustment and impact reduction upon foot touchdown.

The first layer of the foot length control focuses on regulating the robot's leg length during walking. It utilizes contact force feedback to dynamically modify the ankle's vertical displacement, effectively adapting to varying ground heights. However, our experimental findings revealed that this layer alone is insufficient to fully mitigate impact forces during touchdown.

To address this limitation, a second control layer is implemented. This layer leverages the mean value of the bump sensor readings, which are strategically placed on the foot sole, to actively control the vertical distance between the foot and the ground before touchdown and minimize impact forces.

Furthermore, to enhance the robot's adaptability to diverse terrain profiles, the foot compliance control system utilizes the bump sensor values to dynamically adjust the ankle joint's roll and pitch angles. By adjusting these angles, the foot can conform to uneven surfaces, maintaining consistent contact and preventing potential slips.
\subsubsection{Optimization-Based Push Recovery Controller}
In our humanoid robot SURENA-V we employed two fundamental strategies for push recovery while walking: modulation of the Zero Moment Point (ZMP, ankle strategy) and the determination of where and when to take a step (step and step timing adjustment). We applied the optimization-based step control algorithm that utilizes DCM error dynamics \cite{taheri2023robust}.
\begin{itemize}
      \item ZMP Modification Controller:
\end{itemize}
Surena-V has feet, so the ZMP can be modified in the foot margin to eliminate the DCM end-of-step error. The amount of ZMP modification based on the current DCM error ($\pmb{\xi}_{t, err}$) can be obtained according to (\ref{eq:eq1}), and the constraint on this ZMP adjustment is represented as (\ref{eq:eq2}):
\begin{equation}
\pmb{p}_{t, mod}=\frac{\pmb{\xi}_{t, err}}{1-e^{\omega(t-T)}}
\label{eq:eq1}
\end{equation}

\begin{equation}
-\frac{{L}^{foot}}{2} \leqslant {p}_{t, mod,x} \leqslant \frac{{L}^{foot}}{2}
,
-\frac{{W}^{foot}}{2} \leqslant {p}_{t, mod,y} \leqslant \frac{{W}^{foot}}{2}
\label{eq:eq2}
\end{equation}
In (\ref{eq:eq1}) T is the current step-time, and ${L}^{foot}$ and ${W}^{foot}$ are the length and width of the robot foot. The ZMP modification within the foot size can only handle minor disturbances, while more significant disturbances will trigger a stepping strategy.
\begin{itemize}
      \item Step Time and Step Position Controller:
\end{itemize}
In the stepping strategy, the next step location and time are calculated through the quadratic programming (QP) optimisation approach. The DCM error dynamic considering step position adjustment, step time adaptation, and ZMP modification is expressed as follows:
\begin{equation}
\begin{split}
    \pmb{p}_{T_{new}, err} + \pmb{b}_{err} + \pmb{\xi}_{t, ref}e^{-\omega t}(\tau_{ref}-\tau_{new})\\
    = \pmb{p}_{t, mod} + e^{-\omega t}\tau_{new}(\pmb{\xi}_{t,err} - \pmb{p}_{t, mod})
\end{split}
\label{eq:eq3}
\end{equation}
Equation (\ref{eq:eq3}) is the equality constraint of the QP. According to this equation, $\pmb{p}_{t, mod}$ will first recover the current DCM error in a disturbance situation. The additional error that the ZMP modification cannot recover is converted to the change of the next step position ($\pmb{p}_{T_{new}, err}$), next step time ($\pmb\tau_{ref}-\tau_{new}$), and DCM offset ($\pmb{b}_{err}$). These are the sets of optimisation variables that the cost function generates to reduce the errors from the desired step time and pre-planned footprints.
\subsection{Whole-Body Collaboration Control}
\subsubsection{ZMP Modification Using Hand and CoM Motion}
In the context of whole-body collaboration control, the humanoid robot employs an approach to modify its Zero Moment Point (ZMP) through coordinated hand and Center of Mass (CoM) movements. When the robot detects a pressure variation on its fingertip, it initiates hand motion to alleviate the pressure, with the direction of movement determined by comparing pressure levels across different fingers. As the robot's hand reaches its maximum operational range and pressure persists, it strategically adjusts its CoM in the opposite direction of the applied force to maintain stability.
To dynamically regulate the CoM based on measured ZMP and CoM data, the following equation is utilized:
\begin{equation}
\pmb{u} =  -\pmb{k_p}.(\pmb{p}_{zmp, ref} -\pmb{p}_{zmp, m}) + \pmb{k_c}.(\pmb{x}_{com, ref}-\pmb{x}_{com, m})
\label{eq:eq14}
\end{equation}
where $\pmb{u}$ represents the modification in CoM, $\pmb{p}_{zmp, m}$ and $\pmb{p}_{zmp, ref}$ denote measured and desired ZMP values respectively, and $\pmb{x}_{com, m}$ and $\pmb{x}_{com, ref}$ signify measured and desired CoM positions and $\pmb{k_p}$ and $\pmb{k_c}$ are controller gains.
\subsubsection{Step position Adaptation}
The additional current DCM error that cannot be recovered by ZMP modification will activate step position adaptation. Our optimisation problem is configured as follows:
\begin{equation}
\begin{split}
    {w}_{{p,x}}\left \| {p}_{T_{new}, err,x} \right \|^2 + {w}_{{p,y}}\left \| {p}_{T_{new}, err,y} \right \|^2 + {w}_{b,x}\left \| {b}_{err,x} \right \|^2 \\
    + {w}_{b,y}\left \| {b}_{err,y} \right \|^2
\end{split}
\label{eq:eq4}
\end{equation}
In (\ref{eq:eq4}), ${w}_{p,i}$ and ${w}_{b,i}$, $i = x, y$ are weighting factors that are defined according to the importance of each strategy. The kinematic range of the robot leg will limit the amount of ${p}_{T_{new}, err}$ so the next step position can be determined in the following range:
\begin{equation}
-\frac{{L}_{min}^{kin}}{2} \leqslant {p}_{T_{new}, err,x} + {p}_{T_{new}, ref,x} \leqslant \frac{{L}_{max}^{kin}}{2}
\label{eq:eq26}
\end{equation}
\begin{equation}
\frac{{W}_{min}^{kin}}{2} \leqslant {p}_{T_{new}, err,y} + {p}_{T_{new}, ref,y} \leqslant \frac{{W}_{max}^{kin}}{2}
\label{eq:eq5}
\end{equation}
In the sagittal direction, the greatest gap between two feet is denoted as ${L}_{max}^{kin}$. The maximum and minimum distances in lateral direction between two feet are denoted by ${W}_{max}^{kin}$ and ${W}_{min}^{kin}$, respectively.
\section{Experiments and Results}

\begin{figure*}%
    \subfloat[\centering Experiment Setup]{{\includegraphics[trim={0cm 0cm 0cm 0cm}, width=0.35\textwidth]{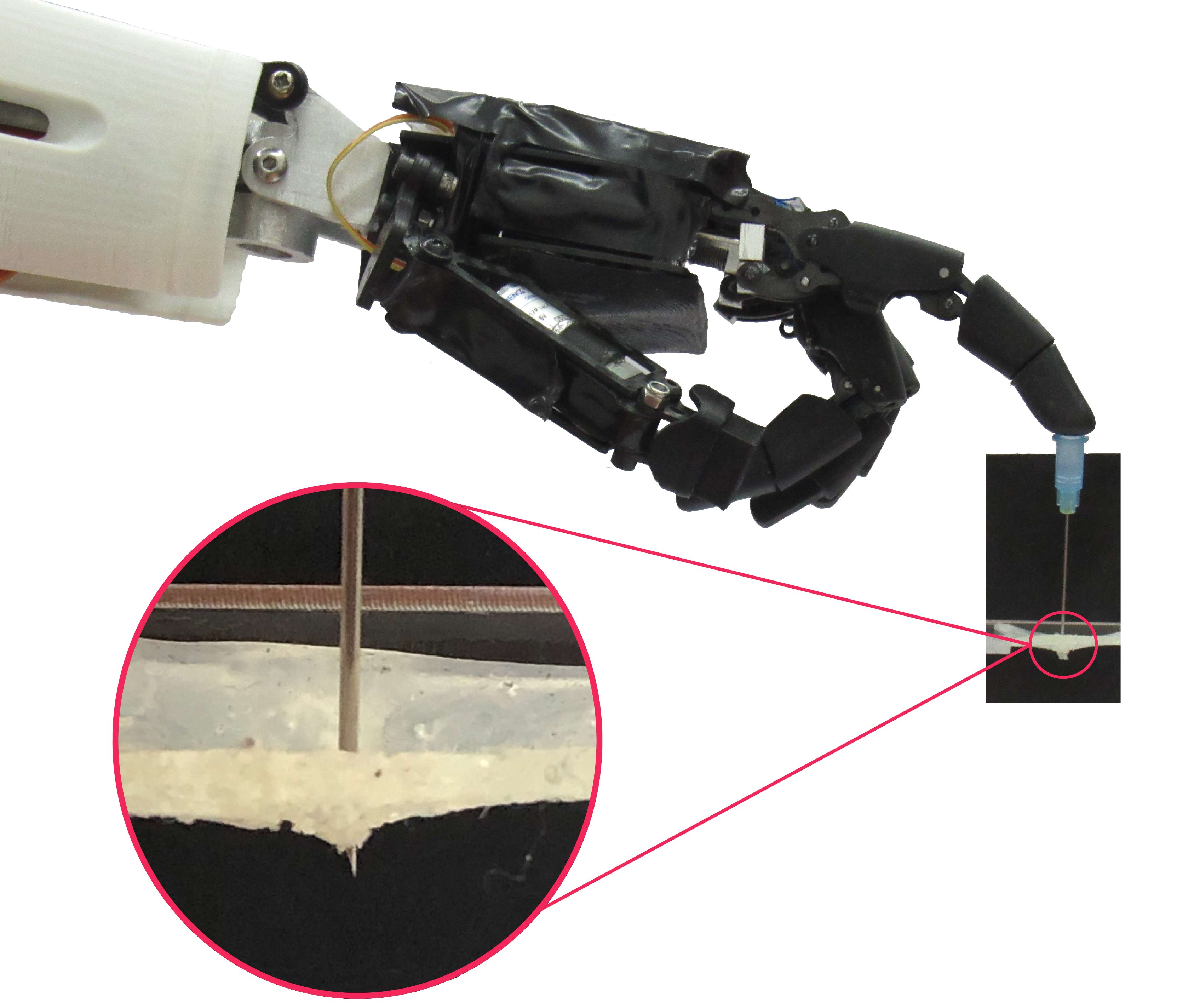} }}%
    \qquad
    \subfloat[\centering Pressure Data]{{\includegraphics[trim={0cm 0cm 0cm 0cm},width=0.55\textwidth]{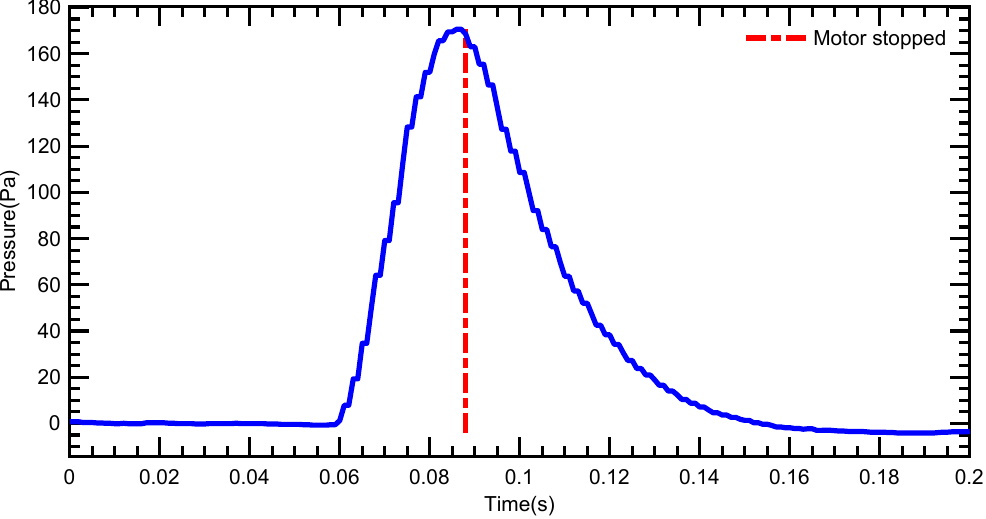} }}%
    \caption{Pressure sensor sensitivity evaluation during pushing a medical needle through a soft material. The robot immediately responds to pressure changes and stops finger movement.}%
    \label{fig:needle_experiment}%
\end{figure*}

\subsection{Sensor Sensitivity}
In this experiment, the sensor sensitivity was evaluated by analyzing its response to a delicate task: pushing a medical needle through a soft material. The primary objective was to assess the robot's ability to detect and react to changes in pressure, particularly when the material was being ripped, in order to ensure safe and precise interactions with different objects.

\begin{figure*}[]
\centering
\includegraphics[width=0.95\textwidth, trim={0 0 0 0}]{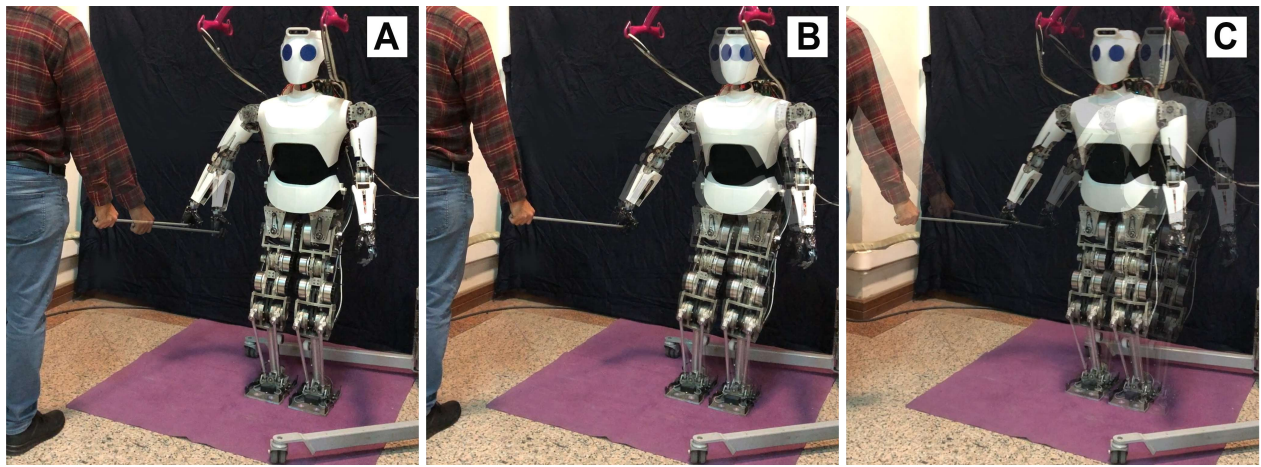}
\label{fig:main_experiment}
\caption{Sequential phases of the collaborative experiment where the humanoid robot interacts with a human participant to manipulate a bar. A: adjusting its arm position, B: Adjusting it's CoM and C: Stepping to right}
\end{figure*}

The sensor utilized in the experiment operates at the update rate of 200 Hz. Furthermore, the sensor exhibits a noise variance of approximately 0.5 Pascal, while still being capable of detecting changes in pressure of less than 1 Pascal with a high degree of accuracy. Figure 4 illustrates the experimental setup and the corresponding pressure data during the experiment. As depicted, the robot ceased finger movement immediately following a decrease in the pressure values.

\subsection{Collaborative Experiment}
In our collaborative experiment, we conducted an evaluation of our designed control architecture by engaging the robot in a task where it collaborates with a human to move a bar. The experiment, depicted in Figure 5, showcases the dynamics of human-robot interaction and the effectiveness of our control strategy.
The robot grasps the bar with a firm grip, utilizing four fingertip pressure sensors to gauge the applied force. Initially, the human participant moves the bar to the left, inducing changes in the pressure sensed by the robot's fingertips. Upon detecting these pressure variations, the robot responds by executing a series of maneuvers to maintain stability and cooperation.
\begin{itemize}
      \item Phase One (Picture A in Figure 5): As the bar is moved to the left, the robot's primary response is to open its arm, restoring the pressure values to their default state.
      \item Phase Two (Picture B in Figure 5): Subsequently, as the bar is moved again, the robot's hand reaches the extremity of its workspace, prompting a shift in the Zero Moment Point (ZMP). This alteration triggers the generation of ZMP control actions, enabling the robot to adjust its Center of Mass (CoM) to counteract potential instability.
      \item Phase Three (Picture C in Figure 5): In the final phase, when the human exerts a strong pull on the bar, the robot dynamically steps to the right to synchronize its movements with the human, ensuring seamless collaboration.
\end{itemize}

\begin{figure*}%
    \subfloat[\centering Little and index fingers pressure data]{{\includegraphics[trim={0cm 0cm 0cm 0cm}, width=0.475\textwidth]{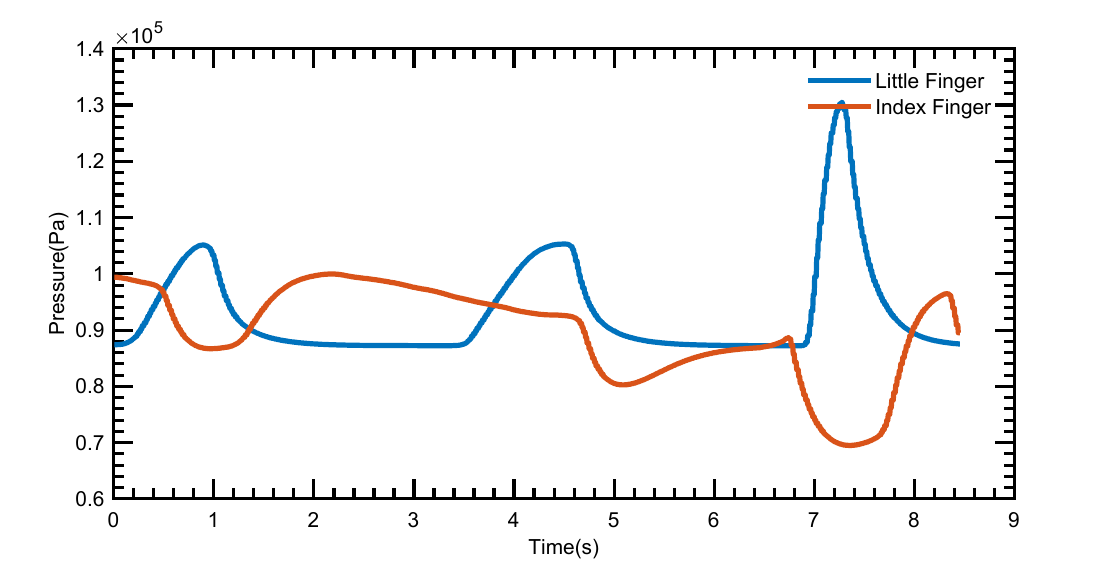} }}%
    \qquad
    \subfloat[\centering Robot key parameters in the y-direction]{{\includegraphics[trim={0cm 0cm 0cm 0cm},width=0.475\textwidth]{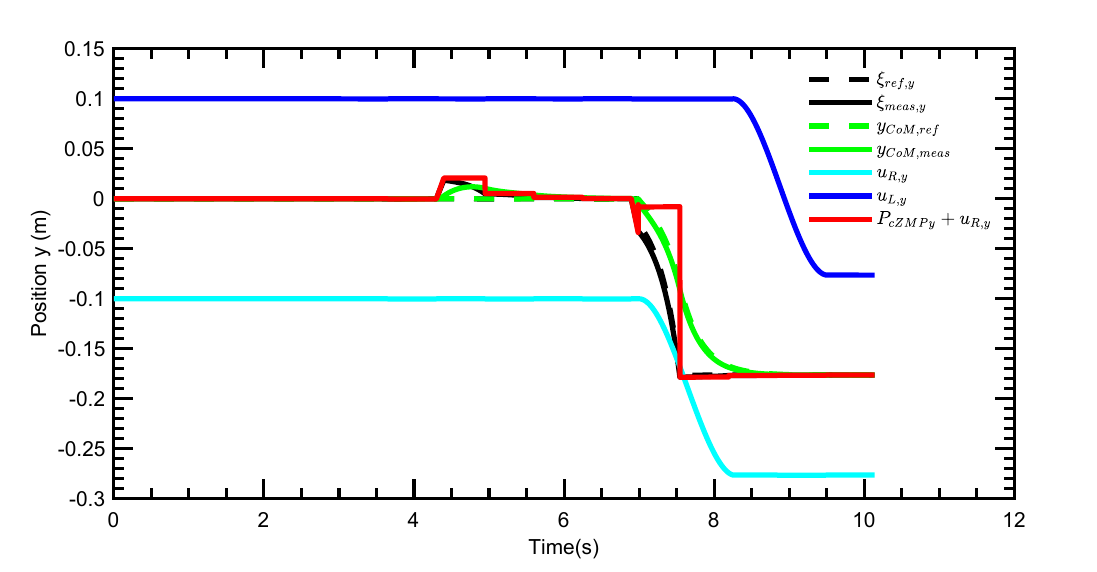} }}%
    \caption{Collaborative experiment results: Between 0 and 2 seconds, the robot opens its arm; between 4 and 6 seconds, it adjusts its CoM; and at 7 seconds, it steps to the right.}%
    \label{fig:needle_experiment}%
\end{figure*}

The pressure readings from the sensors at the little and index fingers, along with the robot's key parameters in y direction, are documented in Figure 6. Between 0 and 2 seconds, the robot detects a change in pressure, prompting it to extend its arms. Subsequently, between 4 and 6 seconds, the robot reaches the end of its workspace, leading to an error in DCM. In response, ZMP modification is generated to adjust the robot's CoM position. By the seventh second, further deviations in DCM error prompt the robot to initiate a lateral step to the right, demonstrating its adaptive response to environmental cues.

\section{Conclusion}
This paper has provided a comprehensive overview of the design and development of the Surena-V humanoid robot. The upper body design incorporated a hand featuring integrated pressure sensors at the fingertips. These sensors play a crucial role in enhancing the robot's interactivity and responsiveness.The pressure sensors' sensitivity was demonstrated by successfully moving a medical needle through soft material. In the lower body design, various mechanisms were employed in the hip, knee, and ankle joints.

The Surena-V humanoid robot showcases an operational architecture that prioritizes stability and collaboration through the integration of various optimization-based control strategies. By detecting and interpreting external forces at their point of effect, such as the robot's hands, Surena-V achieves more agile responses. The robot's ability to seamlessly cooperate with humans in task execution, as demonstrated in the bar-moving experiment, highlights the efficacy of this framework. Future research directions may focus on further enhancing the robot's perceptual capabilities and expanding the range of collaborative tasks it can undertake.


\bibliography{bibliography}
\bibliographystyle{IEEEtran}

\end{document}